\documentclass{article}

\usepackage{algorithm} 
\usepackage{algorithmic} 
\usepackage[utf8]{inputenc} 
\usepackage[T1]{fontenc}    
\usepackage{hyperref}       
\usepackage{url}            
\usepackage{booktabs}       
\usepackage{amsfonts}       
\usepackage{nicefrac}       
\usepackage{microtype}      
\usepackage{graphicx}
\usepackage{amsmath}

\usepackage[svgnames]{xcolor}
\usepackage{xcolor,colortbl} 

\usepackage{caption}
\newfloat{Algorithm}{htbp}{loa}

\usepackage{amssymb}
\usepackage{pifont}
\newcommand{\cmark}{\textcolor{blue}{\ding{51}}}%
\newcommand{\xmark}{\textcolor{red}{\ding{55}}}%

\usepackage[margin= 0.8in,top=12mm]{geometry}

\newcommand*{\titleAT}{\begingroup
  \newlength{\drop}
  \drop=0.05\textheight

  \includegraphics[scale=1.5]{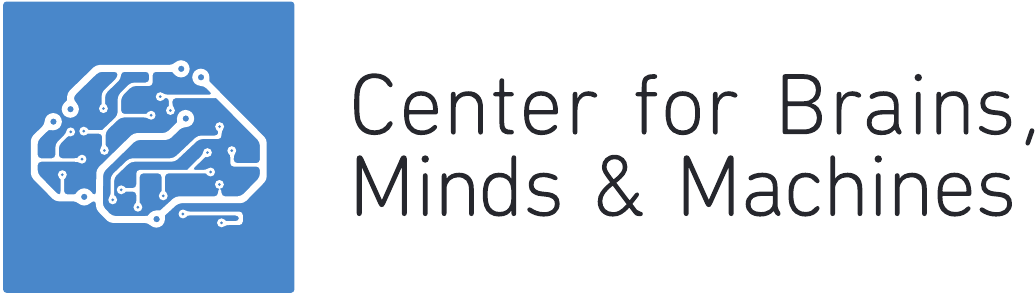}

  \textcolor{CornflowerBlue}{\rule{\textwidth}{3 pt}}\par
  \vspace{2pt}\vspace{-\baselineskip}
  \rule{\textwidth}{0.4pt}\par

  \vspace{\drop}
  \textbf{\large{CBMM Memo No. \memonumber}}   \hfill    \textbf{\large{\memodate}} 

  \vspace{\drop}
  \begin{center}
    \textbf{\huge{\memotitle}}\\
    \vspace{0.4\drop}
    \textbf{\Large{by}}\\
    \vspace{0.4\drop}
    \large{\memoauthors}
  \end{center}
  \vspace{\drop}
  \textbf{\large{\noindent Abstract}:} {\memoabstract}

  \textcolor{CornflowerBlue}{\rule{\textwidth}{3 pt}}\par
  \vspace{2pt}\vspace{-\baselineskip}
  \rule{\textwidth}{0.4pt}\par

  \begin{minipage}{.15\linewidth}
    \includegraphics[scale=0.1]{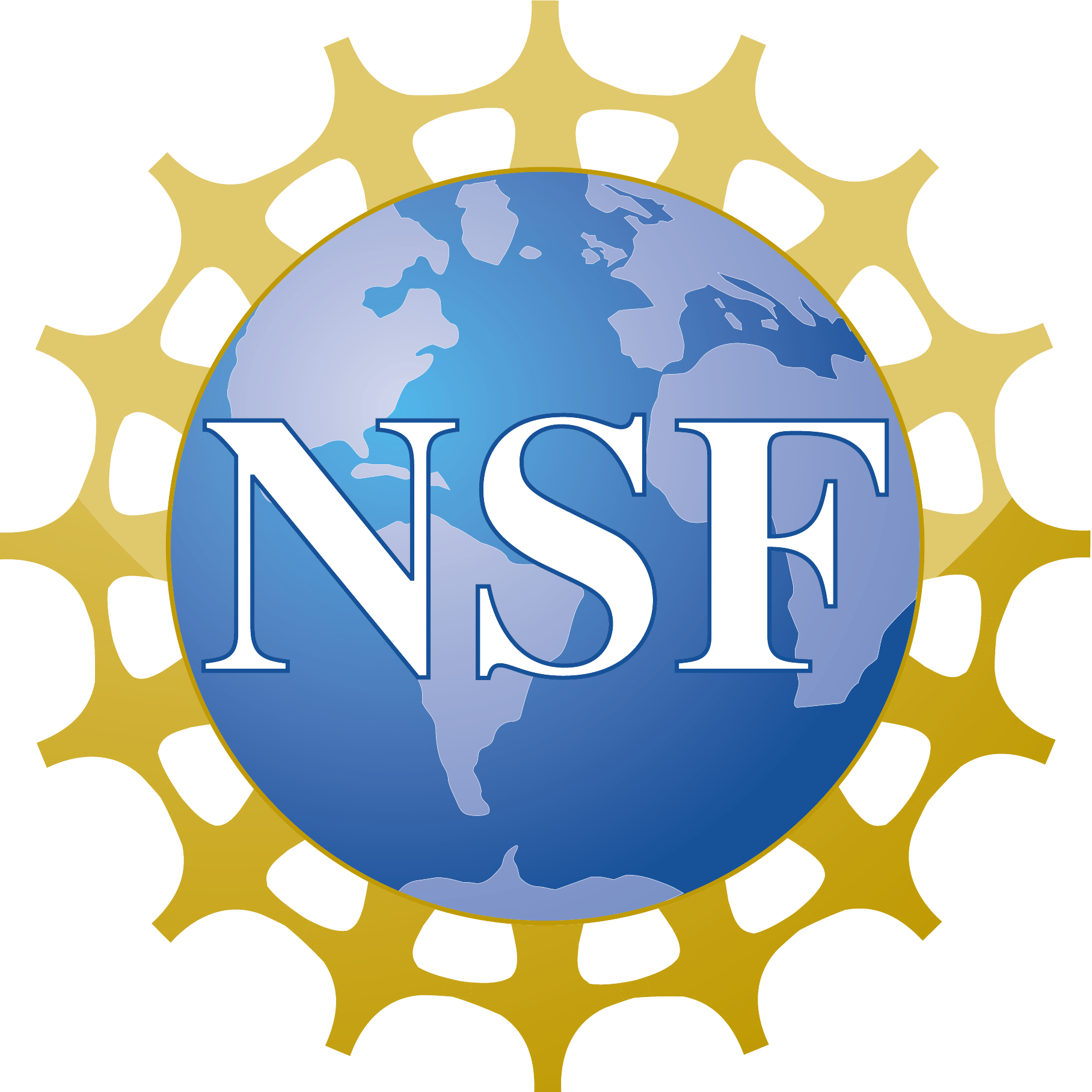}
  \end{minipage}
  \begin{minipage}{.84\linewidth}
    \textbf{\large{This work was supported by the Center for Brains, Minds and Machines (CBMM), funded by NSF STC award  CCF - 1231216.}}
  \end{minipage}
  \endgroup}

\begin{document}


\def\memonumber{057} 
\def\memodate{October 19, 2016}
\def\memotitle{Streaming Normalization: Towards Simpler and More Biologically-plausible Normalizations for Online and Recurrent Learning}
\def\memoauthors{   \textbf{ Qianli Liao$^{1,2}$, Kenji Kawaguchi$^{2}$ and Tomaso Poggio$^{1,2}$ } \\ 1. Center for Brains, Minds and Machines, McGovern Institute \\
  2. Computer Science and Artificial Intelligence Laboratory \\
  MIT
}  


\def\memoabstract{We systematically explored a spectrum of
  normalization algorithms related to Batch Normalization (BN) and
  propose a generalized formulation that simultaneously solves two
  major limitations of BN: (1) online learning and (2) recurrent
  learning. Our proposal is simpler and more biologically-plausible.
  Unlike previous approaches, our technique can be applied out of the
  box to all learning scenarios (e.g., online learning, batch learning,
  fully-connected, convolutional, feedforward, recurrent and mixed ---
  recurrent and convolutional) and compare favorably with existing
  approaches. We also propose Lp Normalization for normalizing by
  different orders of statistical moments. In particular, L1
  normalization is well-performing, simple to implement, fast to
  compute, more biologically-plausible and thus ideal for GPU or
  hardware implementations. }

\titleAT

\newpage

\section{Introduction}

Batch Normalization \cite{ioffe2015batch} (BN) is a highly effective technique
for speeding up convergence of feedforward neural networks. It enabled
recent development of ultra-deep networks \cite{he2015deep} and some
biologically-plausible variants of backpropagation
\cite{liao2015important}. However, despite its success, there are two
major learning scenarios that cannot be handled by BN: (1) online
learning and (2) recurrent learning. 
 
For the second scenario recurrent learning, \cite{liao2016bridging}
and \cite{cooijmans2016recurrent} independently proposed
``time-specific batch normalization'': different normalization
statistics are used for different timesteps of an RNN. Although this
approach works well in many experiments in \cite{liao2016bridging} and
\cite{cooijmans2016recurrent}, it is far from perfect due to the following
reasons: First, it does not work with small mini-batch or online
learning. This is similar to the original batch normalization, where
enough samples are needed to compute good estimates of statistical
moments. Second, it requires sufficient training samples for every timestep of an RNN.
It is not clear how to generalize the model to unseen timesteps.
Finally, it is not biologically-plausible. While homeostatic
plasticity mechanisms (e.g., Synaptic Scaling)
\cite{Turrigiano2004,stellwagen2006synaptic,turrigiano2008self} are
good biological candidates for BN, it is hard to imagine how such
normalizations can behave differently for each timestep. Recently,
Layer Normalization (LN) \cite{ba2016layer} was introduced to solve
some of these issues. It performs well in feedforward and recurrent
settings when only fully-connected layers are used. However, it does
not work well with convolutional networks. A summary of normalization
approaches in various learning scenarios is shown in Table \ref{tab:overview}.

\newcommand{\specialcell}[2][c]{%
  \begin{tabular}[#1]{@{}c@{}}#2\end{tabular}}

\begin{table} 
  \begin{tabular}{  |c|c|c|c|c|c|c|c| }
    \hline
    Approach  & FF \& FC & FF \& Conv & Rec \& FC &  Rec \& Conv  & \specialcell{Online\\ Learning}  & \specialcell{Small\\ Batch} & \specialcell{All \\ Combined}     \\
    \hline
    \specialcell{Original Batch \\ Normalization(BN)}   & \cmark & \cmark &  \xmark  & \xmark   & \xmark & Suboptimal & \xmark   \\ 
    \hline
    Time-specific BN   & \cmark & \cmark  & Limited & Limited &  \xmark & Suboptimal  & \xmark \\ 
    \hline
    \specialcell{Layer\\ Normalization}   &  \cmark  & \xmark*  &  \cmark & \xmark* & \cmark &  \cmark  & \xmark*  \\
    \hline
    \specialcell{Streaming\\ Normalization}    &  \cmark  & \cmark  &  \cmark & \cmark & \cmark &  \cmark  & \cmark \\
    \hline
  \end{tabular}
  \caption{An overview of normalization techiques for different tasks. \cmark: works well. \xmark: does not work well. FF: Feedforward. Rec: Recurrent. FC: Fully-connected. Conv: convolutional. Limited: time-specific BN requires recording normalization statistics for each timestep and thus may not generalize to novel sequence length. *Layer normalization does not fail on these tasks but perform significantly worse than the best approaches. }  
  \label{tab:overview}
\end{table}


We note that different normalization methods like BN and LN can be
described in the same framework detailed in Section
\ref{sec:framework}. This framework introduces Sample Normalization
and General Batch Normalization (GBN) as generalizations of LN and BN,
respectively. As their names imply, they either collect normalization
statistics from a single sample or from a mini-batch. We explored many
variants of these models in the experiment section.

A natural and biologically-inspired extension of these methods would
be Streaming Normalization: normalization statistics are collected in
an online fashion from all previously seen training samples (and all
timesteps if recurrent). We found numerous advantages associated with
this approach: 1. it naturally supports pure online learning or
learning with small mini-batches. 2. for recurrent learning, it is
more biologically-plausible since a unique set of normalization
statistics is maintained for all timesteps. 3. it performs
well out of the box in all learning scenarios (e.g., online learning,
batch learning, fully-connected, convolutional, feedforward, recurrent
and mixed --- recurrent and convolutional). 4. it offers a new
direction of designing normalization algorithms, since the idea of
maintaining online estimates of normalization statistics is
independent from other design choices, and as a result any existing algorithm
(e.g., in Figure \ref{fig:general_norm} C and D) can be extended to a
``streaming'' setting. 


We also propose Lp normalization: instead of normalizing by
the second moment like BN, one can normalize by the p-th root of the p-th
absolute moment (See Section \ref{sec:lp} for details). In particular, L1
normalization works as well as the conventional approach (i.e., L2) in
almost all learning scenarios. Furthermore, L1 normalization is easier to implement: it is
simply the average of absolute values. We believe it can be used to
simplify and speed up BN and our Streaming Normalization in GPGPU,
dedicated hardware or embedded systems. L1 normalization may also be
more biologically-plausible since the gradient of the absolute value is
trivial to implement, even for biological neurons. 

%


In the following section, we introduce a simple training scheme, a minor but necessary component of our formulation.

\section{Online and Batch Learning with ``Decoupled Accumulation and Update''}

Although it is not our main result, we discuss a simple but to the best of our knowledge less explored\footnote{We are not aware of this approach in the literature. If there is, please inform us.} training scheme we call a ``Decoupled Accumulation and Update'' (DAU). 

Conventionally, the weights of a neural network are updated every 
mini-batch. So the accumulation of gradients and weights update are
coupled. We note that a general formulation would be that while for
every mini-batch the gradients are still accumulated, one does not necessarily
update the weights. Instead, the weights are updated every $n$ 
mini-batches. The gradients are cleared after each weight update. 
Two parameters characterize this procedure: Samples per Batch (S/B) $m$ and Batch per Update (B/U) $n$.
In conventional training, $n=1$. 

Note that \textbf{this procedure is similar to but distinct from
  simply training larger mini-batches} since every mini-batch arrives
in a purely online fashion so one cannot look back into any previous
mini-batches. For example, if batch normalization is present,
performing this procedure with $m$ S/B and $n$ B/U is different from that with
$m*n$ S/B and $1$ B/U.

If $m=1$, it reduces to a pure online setting where one sample arrives at a time.
The key advantage of this proposal over conventional training is that
we explicitly require less frequent (but more robust) weight updates. The memory
requirement (in additional to storing the network weights) is storing
$m$ samples and related activations.

This training scheme generalizes the conventional
approach, and we found that merely applying this approach greatly
mitigated (although not completely solved) the catastrophic failure of
training batch normalization with small mini-batches (See Figure \ref{fig:dau}). Therefore, we
will use this formulation throughout our paper.


We also expect this scheme to benefit learning sequential recurrent
networks with varying input sequence lengths. Sometimes it is more
efficient to pack in a mini-batch training samples with the same
sequence length. If this is the case, our approach predicts that it
would be desirable to process multiple such mini-batches with varying 
sequence lengths before a weight update. Accumulating gradients from
training different sequence lengths should provide a more robust update 
that works for different sequence lengths, thus better approximating
the true gradient of the dataset. 

In Streaming Normalization with recurrent networks, it is often
beneficial to learn with $n > 1$ (i.e., more than one batch per
update). The first mini-batch collects all the normalization
statistics from all timesteps so that later mini-batches are
normalized in a more stable way.


\section{A General Framework for Normalization}
\label{sec:framework} 
We propose a general framework to describe different normalization
algorithms. A normalization can be thought of as a process of
modifying the activation of a neuron using some statistics collected
from some reference activations. We adopt three terms to characterize this
process: A \textbf{Normalization Operation (NormOP)} is a function
$N(x_i,s_i)$ that is applied to each neuron $i$ to modify its value
from $x_i$ to $N(x_i,s_i)$, where $s_i$ is the \textbf{Normalization
  Statistics (NormStats)} for this neuron. \textbf{NormStats} is any
data required to perform NormOP, collected using some function $s_i =
S(R_i)$, where $R_i$ is a set of activations (could include $x_i$) called
\textbf{Normalization Reference (NormRef)}. Different neurons may or
may not share NormRef and NormStats. 

\begin{figure}
  \renewcommand\figurename{\small Figure}  
  \begin{center}
    \includegraphics[width=\linewidth]{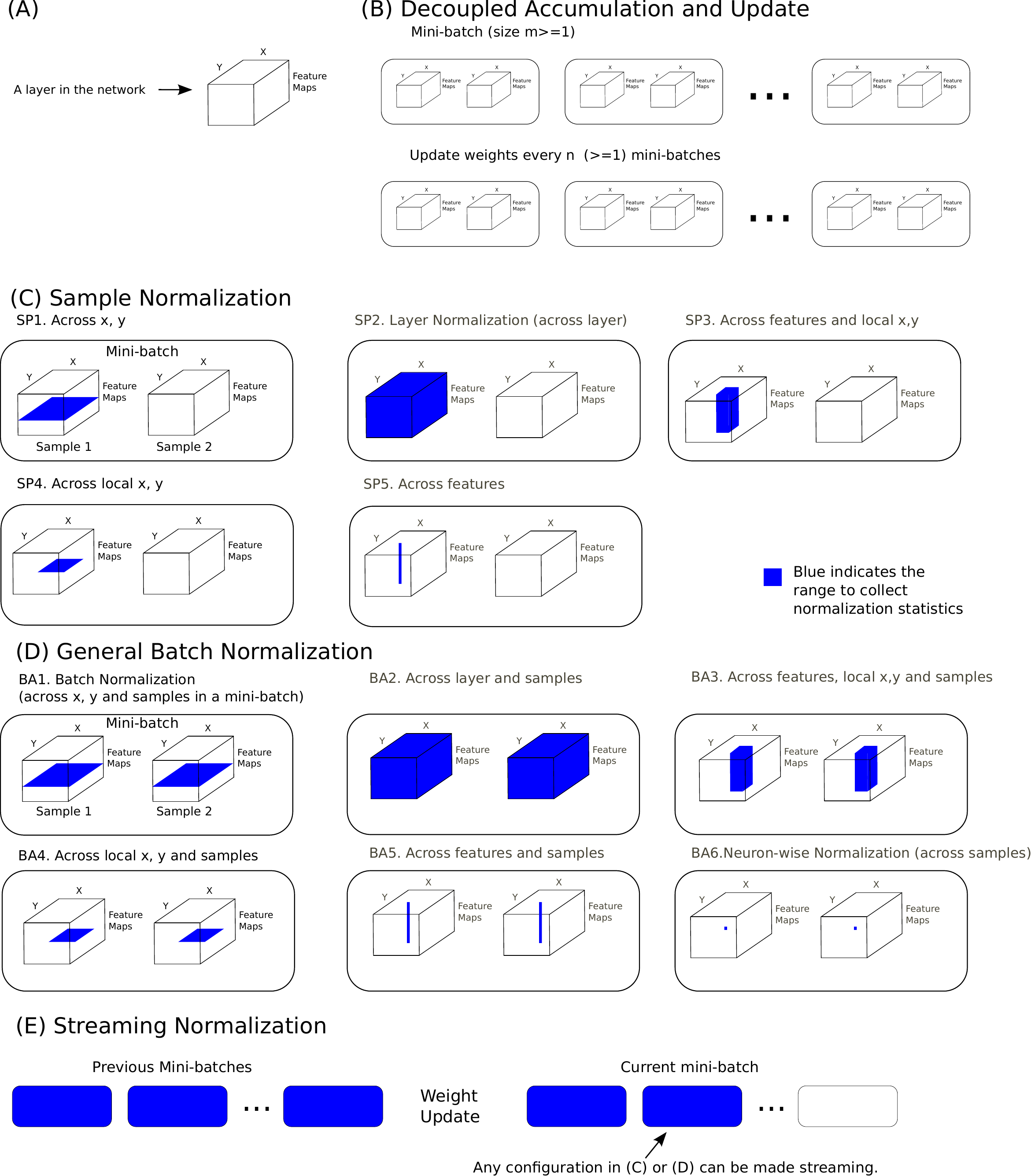}
  \end{center}
  \caption{A General Framework of Normalization. A:  the input to convolutional layer is a 3D matrix consists of 3 dimensions: x (image width), y (image height) and features/channel. For fully-connected layers, x=y=1. B: training with decoupled accumulation and update. C: Sample Normalization. D: General Batch Normalization. E: Streaming Normalization} 
  \label{fig:general_norm}
\end{figure}

This framework captures many previous normalizations algorithms as
special cases. For example, the original Batch Normalization (BN)
\cite{ioffe2015batch} can be described as follows: the NormRef for 
each neuron $i$ is all activations in the same channel of the same
layer and same batch (See Figure \ref{fig:general_norm} C for
illustration). The NormStats are $S_i$ =\{$\mu_i$,$\sigma_i$\} --- the
mean and standard deviation of the activations in NormRef. The NormOp
is N($x_i$,\{$\mu_i$,$\sigma_i$\}) = $\frac{x_i - \mu_i}{\sigma_i}$

For the recent Layer Normalization \cite{ba2016layer}, NormRef is
all activations in the same layer (See Figure
\ref{fig:general_norm} C). The NormStats and NormOp are the same as
BN.



We group normalization algorithms into three categories:
\begin{enumerate}
 \item Sample Normalization (Figure \ref{fig:general_norm} C): NormStats are collected from one sample.
 \item General Batch Normalization (Figure \ref{fig:general_norm} D): NormStats are collected from all samples in a mini-batch. 
 \item Streaming Normalization: NormStats are collected in an online fashion from all pass training samples.
\end{enumerate}
 
In the following sections, we detail each algorithm and provide pseudocode. 


\subsection{Sample Normalization}

Sample normalization is the simplest category among the three.
NormStats are collected only using the activations in the
current layer of current sample. The computation is the same at
training and test times. It handles online learning naturally.
Layer Normalization \cite{ba2016layer} is an example of this category. 
More examples are shown in Figure \ref{fig:general_norm} C.

The pseudocode of forward and backpropagation is show in Algorithm \ref{alg:sample_norm_fprop} and Algorithm \ref{alg:sample_norm_bprop}.  


\begin{algorithm}[H]
  \caption{Sample Normalization Layer: Forward}
  \label{alg:sample_norm_fprop}                           
  \begin{algorithmic}                    
    \REQUIRE layer input \textbf{x} (a sample), NormOP N(.,.), function S(.) to compute NormStats for every element of \textbf{x} 
    \ENSURE layer output \textbf{y} (a sample)
    \STATE s = S(x)
    \STATE y = N(x,s)
  \end{algorithmic} 
\end{algorithm}


\begin{algorithm}[H]
  \caption{Sample Normalization Layer: Backpropagation} 
  \label{alg:sample_norm_bprop}                           
  \begin{algorithmic}                    
    \REQUIRE $\frac{\partial E}{\partial y}$ (a sample) where $E$ is objective, layer input \textbf{x}, NormOP N(.,.), function S(.) to compute NormStats for every element of \textbf{x} 
    \ENSURE $\frac{\partial E}{\partial x}$  (a sample)
    \STATE  $\frac{\partial E}{\partial x}$ can be calculated using chain rule. Detail omitted.
  \end{algorithmic}
\end{algorithm}

\subsection{General Batch Normalization}

In General Batch Normalization (GBN), NormStats are collected in some way using the activations from all samples in a training mini-batch.
Note that one cannot really compute the batch NormStats at test time since
test samples should be handled independently from each
other, instead in a batch. To overcome this, one can simply compute running estimates
of training NormStats and use them for testing (e.g., the original Batch Normalization \cite{ioffe2015batch} computes moving averages).

More examples of GBN are shown in Figure \ref{fig:general_norm} D. The pseudocode is shown in Algorithm \ref{alg:GBN_fprop} and \ref{alg:GBN_bprop}.     



\begin{algorithm}
  \caption{General Batch Normalization Layer: Forward}                           
  \label{alg:GBN_fprop}    
  \begin{algorithmic}                    
    \REQUIRE layer input \textbf{x} (a mini-batch), NormOP N(.,.), function S(.) to compute NormStats for every element of \textbf{x},  running estimates of NormStats \^{s}, function F to update \^{s}.
    \ENSURE layer output \textbf{y} (a mini-batch), running estimates of NormStats \^{s}
    \IF {$training$}
    \STATE s = S(x)
    \STATE \^{s} = F(\^{s},s)
    \STATE y = N(x,s)
    \ELSE [$testing$] 
    \STATE y = N(x,\^{s}) 
    \ENDIF 
  \end{algorithmic} 
\end{algorithm}

\begin{algorithm}[H]
  \caption{General Batch Normalization Layer: Backpropagation}
  \label{alg:GBN_bprop}                           
  \begin{algorithmic}                    
    \REQUIRE $\frac{\partial E}{\partial y}$ (a mini-batch) where $E$ is objective, layer input \textbf{x} (a mini-batch), NormOP N(.,.), function S(.) to compute NormStats for every element of \textbf{x},  running estimates of NormStats \^{s}, function F to update \^{s}.
    \ENSURE $\frac{\partial E}{\partial x}$  (a mini-batch) 
    \STATE  $\frac{\partial E}{\partial x}$ can be calculated using chain rule. Detail omitted.
  \end{algorithmic}
\end{algorithm}






\subsection{Streaming Normalization} 

Finally, we present the main results of this paper. We propose
Streaming Normalization: NormStats are collected in an online fashion
from all past training samples. The main challenge of this approach is
that it introduces infinitely long dependencies throughout the
neuron's activation history --- every neuron's current activation
depends on all previously seen training samples.

It is intractable to perform exact backpropagation on this dependency
graph for the following reasons: there is no point backpropagating
beyond the last weight update (if any) since one cannot redo the
weight update. This, on the other hand, would imply that one cannot update the weights until having
seen all future samples. Even backpropagating within a weight update
(i.e., the interval between two weight updates, consisting of $n$
mini-batches) turns out to be problematic: one is usually not allowed
to backpropagate to the previous several mini-batches since they are
discarded in many practical settings.  

For the above reasons, we abandoned the idea of performing exact backpropagation for streaming normalization.
Instead, we propose two simple heuristics: Streaming NormStats and Streaming Gradients.
They are discussed below and the pseudocode is shown in Algorithm \ref{alg:SN_fprop} and Algorithm \ref{alg:SN_bprop}.

\subsubsection{Streaming NormStats} 
\label{sec:stream_stats}  
Streaming NormStats is a natural 
requirement of Streaming Normalization, since NormStats are 
collected from all previously seen training samples. We maintain a
structure/table $H_1$ to keep all the information needed to generate a
good estimate of NormStats. and update it using a function $F$
everytime we encounter a new training sample. Function $F$ also
generates the current estimate of NormStats called $\hat{s}$. We use
$\hat{s}$ to normalize instead of $s$. See Algorithm
\ref{alg:SN_fprop} for details.

There could be many potential designs for $F$ and $H_1$, and in our experiments we explored a particular version: we compute two sets
of running estimates to keep track of the long-term and short-term 
NormStats: $H_1 = \{ \hat{s}_{long}, \hat{s}_{short}, counter \}$.

\begin{itemize}
\item Short-term NormStats $\hat{s}_{short}$ is the \textbf{exact
  average} of NormStats $s$ since \textbf{the last weight update}.
  The $counter$ keeps track of the number of times a different $s$ is encountered
  to compute the exact average of $s$.
\item Long-term NormStats $\hat{s}_{long}$ is an \textbf{exponential
  average} of $\hat{s}_{short}$ since \textbf{the beginning of
  training}.
\end{itemize}

Whenever the weights of the network is updated: 
$\hat{s}_{long} = \kappa_1 * \hat{s}_{long} + \kappa_2 *
\hat{s}_{short}$, $counter$ is reset to 0 and $\hat{s}_{short}$ is set
to empty. In our experiments, $\kappa_1+\kappa_2=1$ so that an exponential average of $\hat{s}_{short}$ is maintained in $\hat{s}_{long}$.

In our implementation, before testing the model, the last weight
update is NOT performed, and $\hat{s}_{short}$ is also NOT cleared,
since $\hat{s}_{short}$ is needed for testing \footnote{It also
possible to simply store the last $\hat{s}$ computed in training for
testing, instead of storing $\hat{s}_{short}$ and re-computing $\hat{s}$ in testing. These two options are mathematically equivalent.} 

In addition to updating $H_1$ in the way described above, the function $F$ also computes
$\hat{s}=\alpha_1\hat{s}_{long} +
\alpha_2\hat{s}_{short}$. Finally, $\hat{s}$ is used for
normalization.




\subsubsection{Streaming Gradients}
\label{sec:stream_grads} 
We maintain a structure/table $H_2$ to
keep all the information needed to generate a good estimate of
gradients of NormStats and update it using a function $G$ everytime
backpropagation reaches this layer. Function $G$ also generates the
current estimate of gradients of NormStats called
$\widehat{\frac{\partial E}{\partial \hat{s}}}$. We use
$\widehat{\frac{\partial E}{\partial \hat{s}}}$ for further
backpropagation instead of $\frac{\partial E}{\partial \hat{s}}$. See
Algorithm \ref{alg:SN_bprop} for details.

Again, there could be many potential designs for $G$ and $H_2$, and we
explored a particular version: we compute two sets
of running estimates to keep track of the long-term and short-term
gradients of NormStats: $H_2 = \{ \hat{g}_{long}, \hat{g}_{short}, counter \}$.  

\begin{itemize}
\item Short-term Gradients of NormStats $\hat{g}_{short}$ is the \textbf{exact average}
  of gradients of NormStats $\frac{\partial E}{\partial \hat{s}}$ since \textbf{the last weight update}.
The $counter$ keeps track of the number of times a different  $\frac{\partial E}{\partial \hat{s}}$ is encountered
  to compute the exact average of  $\frac{\partial E}{\partial \hat{s}}$.
\item Long-term Gradients of NormStats $\hat{g}_{long}$ is an \textbf{exponential
  average} of $\hat{g}_{short}$ since \textbf{the beginning of
  training}.
\end{itemize}

Whenever the weights of network is updated:
$\hat{g}_{long} = \kappa_3 * \hat{g}_{long} + \kappa_4 *
\hat{g}_{short}$, $counter$ is reset to 0 and $\hat{g}_{short}$ is set
to empty. In our experiments, $\kappa_3+\kappa_4=1$ so that an exponential average of $\hat{g}_{short}$ is maintained in $\hat{g}_{long}$. 

In addition to updating $H_2$ in the way described above, the function $G$ also computes
$\widehat{\frac{\partial E}{\partial \hat{s}}}=\beta_1\hat{g}_{long} +
\beta_2\hat{g}_{short} + \beta_3 \frac{\partial E}{\partial \hat{s}}$. Finally, $\widehat{\frac{\partial E}{\partial \hat{s}}}$ is used for 
further backpropagation.

\begin{algorithm}
  \begin{algorithmic}                    
    \REQUIRE layer input \textbf{x} (a mini-batch), NormOP N(.,.), function S(.) to compute NormStats for every element of \textbf{x},  running estimates of NormStats and/or related information packed in a structure/table $H_1$, function F to update $H_1$ and generate current estimate of NormStats \^{s}. 
    \ENSURE layer output \textbf{y} (a mini-batch) and $H_1$ (it is stored in this layer, instead of feeding to other layers), always maintain the latest \^{s} in case of testing
    \IF {$training$}
    \STATE s = S(x) 
    \STATE \{$H_1$,\^{s}\} = F($H_1$,s) 
    \STATE y = N(x,\^{s})
    \ELSE [$testing$] 
    \STATE y = N(x,\^{s})
    \ENDIF 
  \end{algorithmic} 
  \caption{Streaming Normalization Layer: Forward}  
  \label{alg:SN_fprop} 
\end{algorithm}

\begin{algorithm}
  \begin{algorithmic}                    
    \REQUIRE $\frac{\partial E}{\partial y}$ (a mini-batch) where $E$ is objective, layer input \textbf{x} (a mini-batch), NormOP N(.,.), function S(.) to compute NormStats for every element of \textbf{x},  running estimates of NormStats \^{s}, running estimates of gradients and/or related information packed in a structure/table $H_2$, function G to update $H_2$ and generate the current estimates of gradients of NormStats $\widehat{\frac{\partial E}{\partial \hat{s}}}$.
    \ENSURE $\frac{\partial E}{\partial x}$  (a mini-batch) and $H_2$  (it is stored in this layer, instead of feeding to other layers)
    \STATE  $\frac{\partial E}{\partial \hat{s}}$ is calculated using chain rule.
    \STATE  \{$H_2$,$\widehat{\frac{\partial E}{\partial \hat{s}}}$\} = G($H_2$,$\frac{\partial E}{\partial \hat{s}}$)  
    \STATE  Use $\widehat{\frac{\partial E}{\partial \hat{s}}}$ for further backpropagation, instead of $\frac{\partial E}{\partial \hat{s}}$
    \STATE  $\frac{\partial E}{\partial x}$ is calculated using chain rule.
  \end{algorithmic}
  \caption{Streaming Normalization Layer: Backpropagation}
  \label{alg:SN_bprop}                           
\end{algorithm}

\subsubsection{A Summary of Streaming Normalization Design and Hyperparameters}
\label{sec:hyper} 

The NormOp used in this paper is N($x_i$,\{$\mu_i$,$\sigma_i$\}) = $\frac{x_i - \mu_i}{\sigma_i}$, the same as what is used by BN. The NormStats are collected using one of the Lp normalization schemes described in Section \ref{sec:lp}.  

With our particular choices of $F$, $G$, $H_1$ and $H_2$, the following hyperparameters uniquely characterize a Streaming Normalization algorithm: $\alpha_1,\alpha_2$, $\beta_1,\beta_2,\beta_3$, $\kappa_1,\kappa_2,\kappa_3,\kappa_4$, samples per batch $m$, batches per update $n$ and a choice of mini-batch NormRef (i.e., SP1-SP5, BA1-BA6 in Figure \ref{fig:general_norm} C and D). 
 
Unless mentioned otherwise, we use BA1 in Figure \ref{fig:general_norm} D as the NormRef throughout the paper. We also demonstrate the use of other NormRefs (BA4 and BA6) in the Appendix Figure \ref{fig:variants}. 

An important special case: If $n=1,\alpha_1=0,\alpha_2=1,\beta_1=0,\beta_2=0,\beta_3=1$, we ignore all NormStats and gradients beyond the current mini-batch. \textbf{The algorithm reduces to exactly the GBN algorithm.} So Streaming Normalization is strictly a generalization of GBN and thus also captures the original BN \cite{ioffe2015batch} as a special case. Recall from Section \ref{sec:stream_stats} that $\hat{s}_{short}$ is NOT cleared before testing the model. Thus for testing, NormStats are inherited from the last training mini-batch. It works well in practice.    

Unless mentioned otherwise, we set $\alpha_1=\kappa_1=\kappa_3$,  $\alpha_2=\kappa_2=\kappa_4$,  $\kappa_1+\kappa_2=1$,  $\kappa_3+\kappa_4=1$,  $\alpha_1+\alpha_2=1$, $\beta_1+\beta_2+\beta_3=1$. We leave it to future research to explore different choices of hyperparameters (and perhaps other $F$ and $G$).

\subsubsection{Implementation Notes}  
One minor drawback of not performing exact backpropagation is that it may break the gradient check of the entire model.
One solution coulde be: (1) perform gradient check of the model without SN and then add a correctly implemented SN. (2) make sure the SN layer is correctly coded (e.g., by reducing it to standard BN using the hyperparameters discussed above and then perform gradient check).

\subsection{Lp Normalization: Calculating NormStats with Different Orders of Moments}
\label{sec:lp}
Let us discuss the function $S(.)$ for calculating NormStats. There are several choices for this function. We propose \textbf{Lp normalization}. It captures the previous mean-and-standard-deviation normalization as a special case. 

First, mean $\mu$ is always calculated the same way --- the average of the activations in NormRef. 
The \textbf{divisive factor $\sigma$}, however, can be calculated in several different ways. In \textbf{Lp Normalization}, $\sigma$ is chosen to be the $p$-th root of the $p$-th \textbf{Absolute Moment}.

Here the \textbf{Absolute Moment} of a distribution $P(x)$ about a point $c$ is:

\begin{equation}
  \int |x-c|^p P(x)dx 
\end{equation}
and the discrete form is:
\begin{equation}
  \frac{1}{N} \sum_{i=1}^{N} |x_i-c|^p
\end{equation}

Lp Normalization can be performed with three settings: 
\begin{itemize}
\item \textbf{Setting A:} $\sigma$ is the $p$-th root of the $p$-th absolute moment of all activations in NormRef with $c$ being the mean $\mu$ of NormRef.
\item \textbf{Setting B:} $\sigma$ is the $p$-th root of the $p$-th absolute moment of all activations in NormRef with $c$ being the running estimate $\hat{\mu}$ of the average.
\item \textbf{Setting C:} $\sigma$ is the $p$-th root of the $p$-th absolute moment of all activations in NormRef with $c$ being 0.
\end{itemize} 
Most of these variants have similar performance but some are better in some situations. 

We call it Lp normalization since it is similar to the norm in the $L^p$ space. 

Setting B and C are better for online learning since A will give degenerate result (i.e., $\sigma=0$) when there is only one sample in NormRef.
Empirically, when there are enough samples in a mini-batch, A and B perform similarly. 

We discuss several important special cases:

\textbf{Special Case A-2: setting A with n=2, $\sigma$ is the standard deviation (square root of the 2nd moment) of all activations in NormRef.} This setting is what is used by Batch Normalization \cite{ioffe2015batch} and Layer Normalization \cite{ba2016layer}. 

\textbf{Special Case p=1:} Whenever p=1, $\sigma$ is simply the average of absolute values of activations. This setting works virtually the same as p=2, but is much simpler to implement and faster to run. It might also be more biologically-plausible, since the gradient computations are much simpler.

\subsection{Separate Learnable Bias and Gain Parameters} 
The original Batch Normalization \cite{ioffe2015batch} also learns a bias and a gain (i.e., shift and scaling) parameter for each feature map. Although not usually done, but clearly these shift and scaling operations can be completely separated from the normalization layer. We implemented them as a separate layer following each normalization layer. These parameters are learned in the same way for all normalization schemes evaluated in this paper. 


\section{Generalization to Recurrent Learning}
In this section, we generalize Sample Normalization, General Batch Normalization and Streaming Normalization to recurrent learning. The difference between recurrent learning and feedforward learning is that for each training sample, every hidden layer of the network receives $t$ activations $h_1,,...,h_t$, instead of only one.


\subsection{Recurrent Sample Normalization}
Sample Normalization naturally generalizes to recurrent learning since all NormStats are collected from the current layer of the current timestep. Training and testing algorithms remain the same.  

\subsection{Recurrent General Batch Normalization (RGBN)}

The generalization of GBN to recurrent learning is the same as what was proposed by \cite{liao2016bridging} and \cite{cooijmans2016recurrent}. The training procedure is exactly the same as before (Algorithm \ref{alg:GBN_fprop} and Algorithm \ref{alg:GBN_bprop}). For testing, one set of running estimates of NormStats is maintained for each timestep. 


Another way of viewing this algorithm is that the same GBN layers described in Algorithm \ref{alg:GBN_fprop} and \ref{alg:GBN_bprop} are used in the \textbf{unrolled} recurrent network. Each GBN layer in the unrolled network uses a different memory storage for NormStats.

\subsection{Recurrent Streaming Normalization}

Extending Streaming Normalization to recurrent learning is
straightforward -- we not only stream through all the past samples,
but also through all past timesteps. Thus, we maintain a unique set of
running estimates of NormStats for all timesteps. This is more
biologically-plausible and memory efficient than the above approach
(RGBN).

Again, another way of viewing this algorithm is that the same Streaming Normalization layers described in Algorithm \ref{alg:SN_fprop} and \ref{alg:SN_bprop} are used in the \textbf{original (instead of the unrolled)} recurrent network. All unrolled versions of the same layer share running estimates of NormStats and other related data. 

One caveat is that as time proceeds, the running estimates of
NormStats are slightly modified. Thus when backpropagation reaches the
same layer again, the NormStats are slightly different from the ones
originally used for normalization. Empirically, it seems to not cause
any problem on the performance. Training with ``decoupled accumulation
update'' with Batches per Update (B/U) > 1 \footnote{B/U=2 is often
  enough} can also mitigate this problem, since it makes NormStats more stable over time. 

\section{Streaming Normalized RNN and GRU}
\label{sec:rnn_gru}
In our character-level language modeling task, we tried Normalized Recurrent Neural Network (RNN) and Normalized Gated Recurrent Unit (GRU) \cite{chung2014empirical}. Let us use Norm(.) to denote a normalization, which can be either Sample Normalization, General Batch Normalization or Streaming Normalization. A bias and gain parameter is also learned for each neuron.  We use NonLinear to denote a nonlinear function. We used hyperbolic tangent (tanh) nonlinearity in our experiments. But we observed ReLU also works. $h_t$ is the hidden activation at time $t$. $x_t$ is the network input at time $t$.  $W_{.}$ denotes the weights. $\odot$ denotes elementwise multiplication.
 
\textbf{Normalized RNN}
\begin{equation}
 h_t = NonLinear(Norm(W_x*x_t) + Norm(W_h*h_{t-1}))
\end{equation}

\textbf{Normalized GRU}
\begin{align}
  g_r = Sigmoid( Norm(W_{xr}*x_t) + Norm(W_{hr}*h_{t-1})) ) \\
  g_z = Sigmoid( Norm(W_{xz}*x_t) + Norm(W_{hz}*h_{t-1})) ) \\
  h_{new}  = NonLinear(Norm(W_{xh}*x_t) + Norm(W_{hh}*(h_{t-1} \odot g_r) )) \\ 
  h_t     = g_z \odot h_{new} + (1 - g_z) \odot h_{t-1}   
\end{align}

\section{Related Work}
\cite{laurent2015batch} and \cite{amodei2015deep} used Batch Normalization (BN) in stacked recurrent networks, where BN was only applied to the feedforward part (i.e., ``vertical'' connections, input to each RNN), but not the recurrent part (i.e., ``horizontal'', hidden-to-hidden connections between timesteps). \cite{liao2016bridging} and \cite{cooijmans2016recurrent} independently proposed applying BN in recurrent/hidden-to-hidden connections of recurrent networks, but separate normalization statistics must be maintained for each timestep.  \cite{liao2016bridging} demonstrated this idea with deep multi-stage fully recurrent (and convolutional) neural networks with ReLU nonlinearities and residual connections. \cite{cooijmans2016recurrent} demonstrated this idea with LSTMs on language processing tasks and sequential MNIST. \cite{ba2016layer} proposed Layer Normalization (LN) as a simple normalization technique for online and recurrent learning. But they observed that LN does not work well with convolutional networks.   \cite{salimans2016weight} and \cite{neyshabur2015path} studied normalization using weight reparameterizations. An early work by \cite{ullman1982adaptation} mathematically analyzed a form of online normalization for visual perception and adaptation.

\section{Experiments}


\subsection{CIFAR-10 architectures and Settings}

\begin{figure}[H]
  \renewcommand\figurename{\small Figure}  
  \begin{center}
    \includegraphics[width=\linewidth]{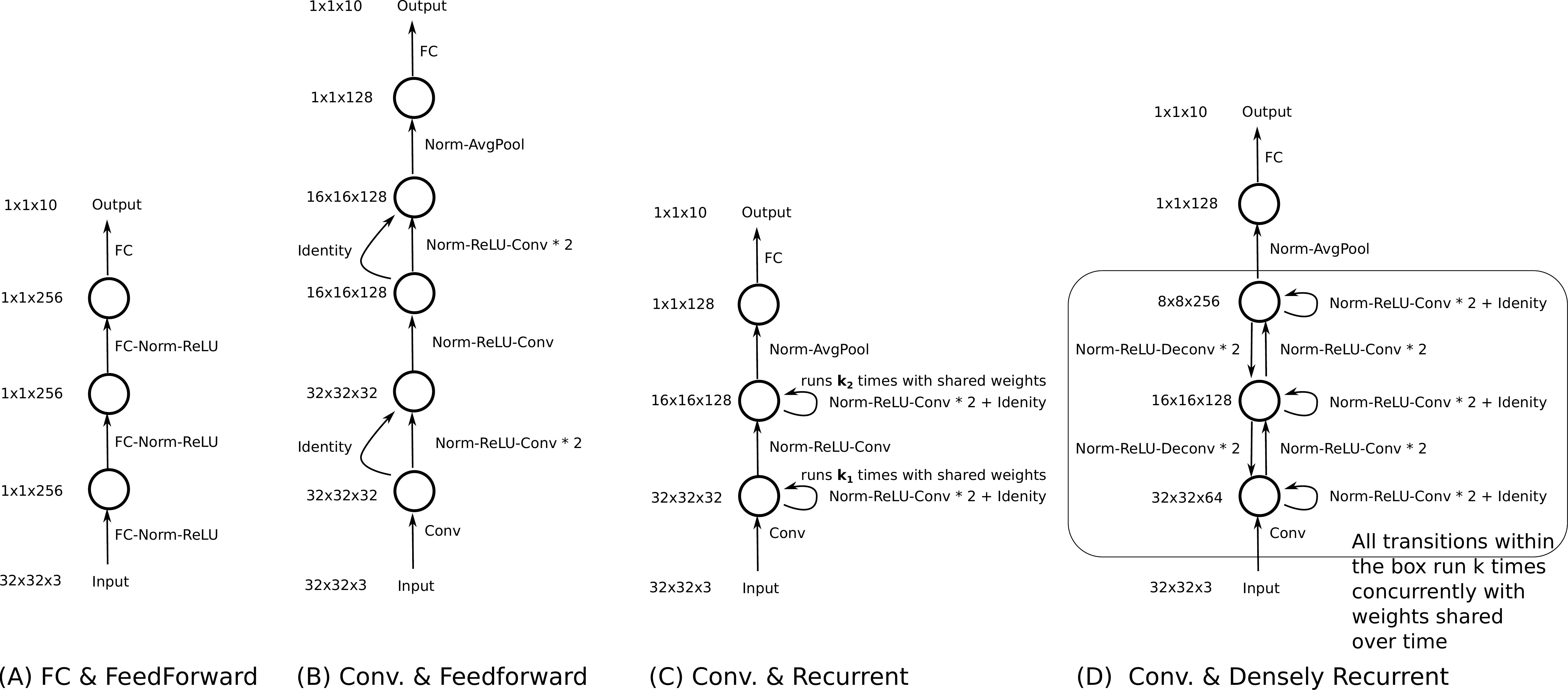}
  \end{center}
  \caption{Architectures for CIFAR-10. Note that C reduces to B when $k_1=k_2=1$.  }    
  \label{fig:cifar_arch} 
\end{figure}

We evaluated the normalization techniques on CIFAR-10 dataset using feedforward fully-connected networks, feedforward convolutional network and a class of convolutional recurrent networks proposed by \cite{liao2016bridging}. The architectural details are shown in Figure \ref{fig:cifar_arch}.  We train all models with learning rate 0.1 for 25 epochs and 0.01 for 5 epochs. Momentum 0.9 is used. We used MatConvNet \cite{vedaldi2015matconvnet} to implement our models.


\subsection{Lp Normalization}

\begin{figure}
  \renewcommand\figurename{\small Figure}  
  \begin{center}
    \includegraphics[width=\linewidth]{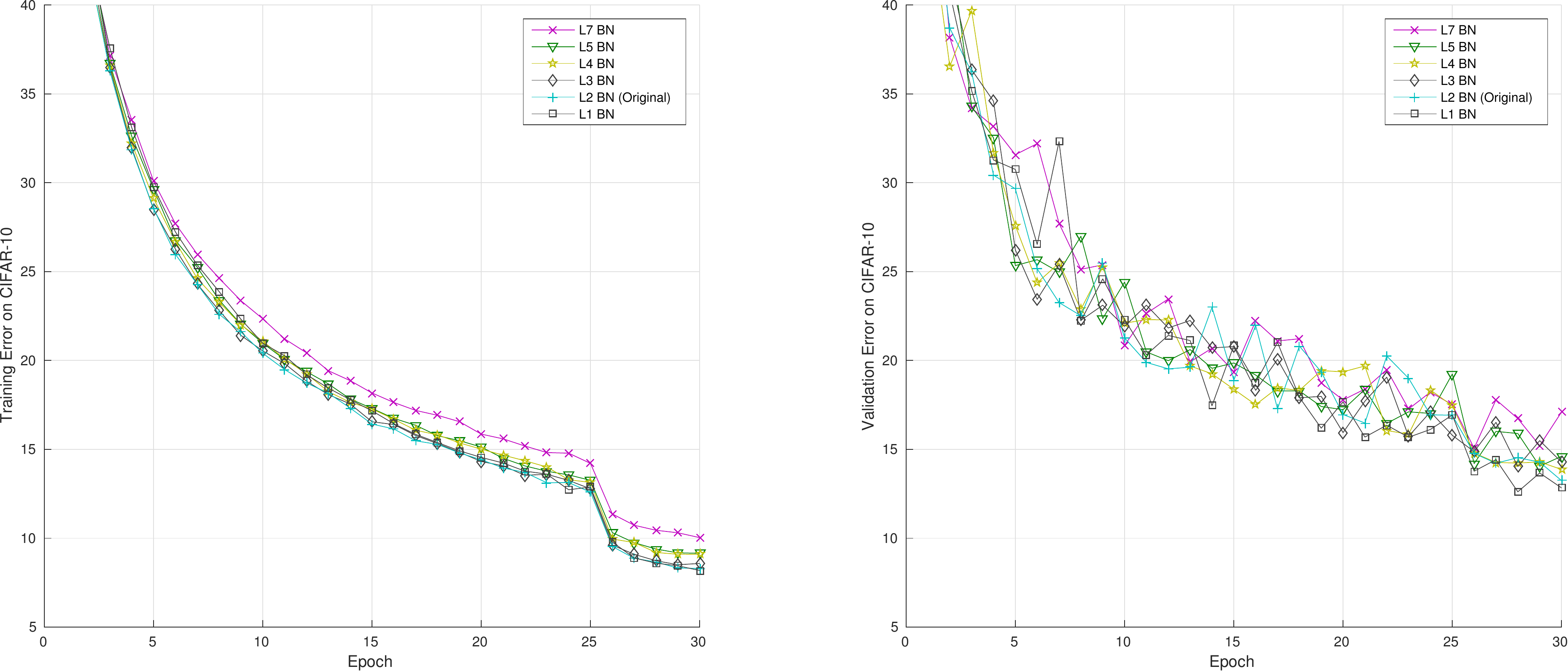}  
  \end{center}
  \caption{Lp Normalization. The architecture is a \textbf{feedforward and convolutional} network (shown in Figure \ref{fig:cifar_arch} B).  All statistical moments perform similarly well. L7 normalization is slightly worse.   } 
  \label{fig:lp_b} 
\end{figure}

We show BN with Lp normalization in Figure \ref{fig:lp_b}. Note that Lp normalization can be applied to Layer Normalization and all other normalizations show in \ref{fig:general_norm} C and D.  L1 normalization works as well as L2 while being simpler to implement and faster to compute. 


\subsection{Online Learning or Learning with Very Small Mini-batches}
We perform online learning or learning with small mini-batches using architecture A in Figure \ref{fig:cifar_arch}.

\begin{figure}[H]
  \renewcommand\figurename{\small Figure}  
  \begin{center}
    \includegraphics[width=0.8\linewidth]{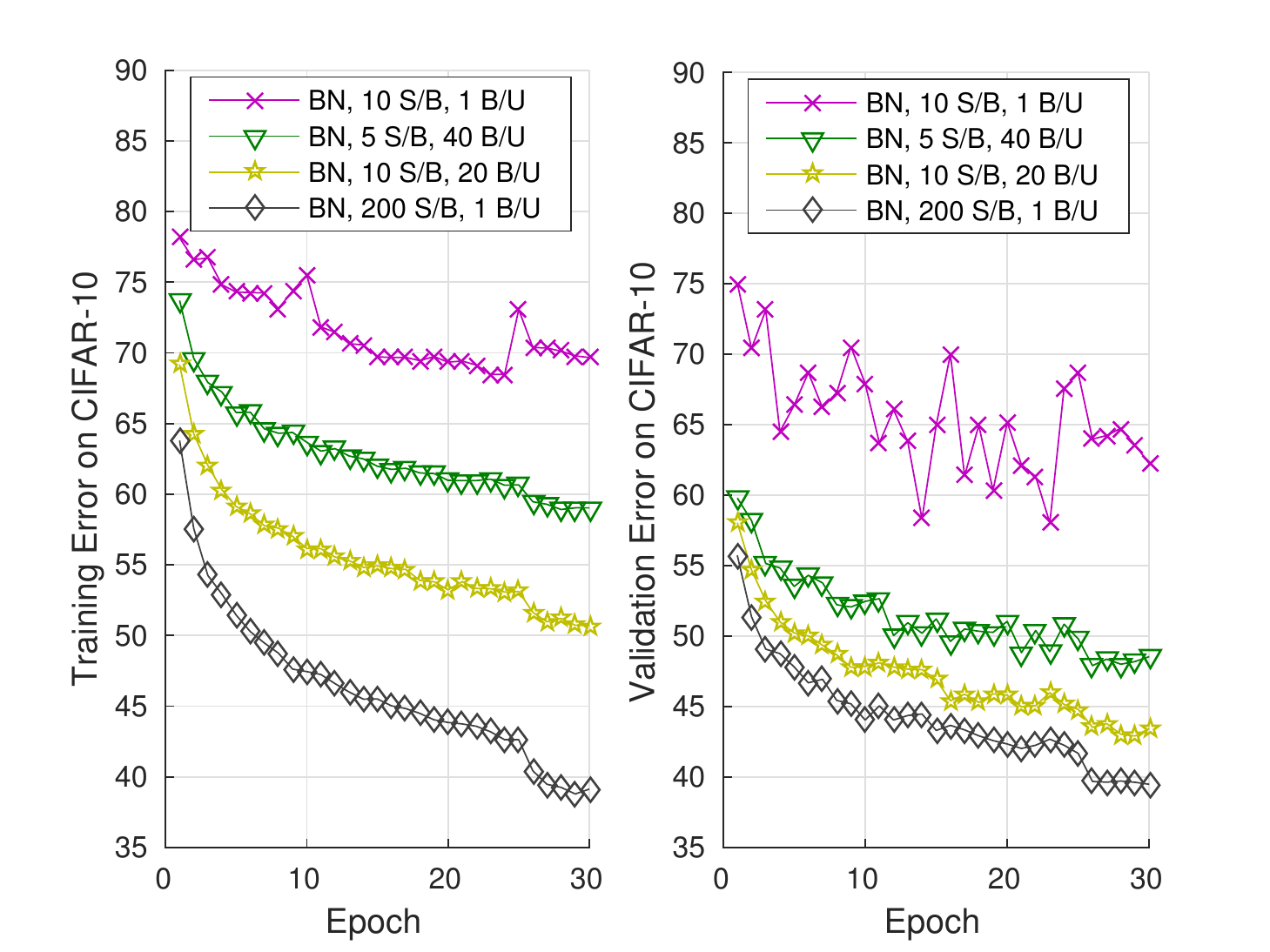}
  \end{center}
  \caption{Plain Mini-batch vs. Decoupled Accumulation and Update (DAU). The architecture is a \textbf{feedforward and fully-connected} network (shown in Figure \ref{fig:cifar_arch} A). S/B: Samples per Batch. B/U: Batches per Weight Update. We show there are significant performance differences between plain mini-batch (i.e., B/U=1) and Decoupled Accumulation and Update (DAU, i.e., B/U=n>1). DAU significantly improves the performance of BN with small number of samples per mini-batch (e.g., compare curve 1 with 3). } 
  \label{fig:dau}
\end{figure}

\textbf{Plain Mini-batch vs. Decoupled Accumulation and Update (DAU): } We show in Figure \ref{fig:dau} comparisons between conventional mini-batch training and Decoupled Accumulation and Update (DAU).

\begin{figure}[H]
  \renewcommand\figurename{\small Figure}  
  \begin{center}
    \includegraphics[width=\linewidth]{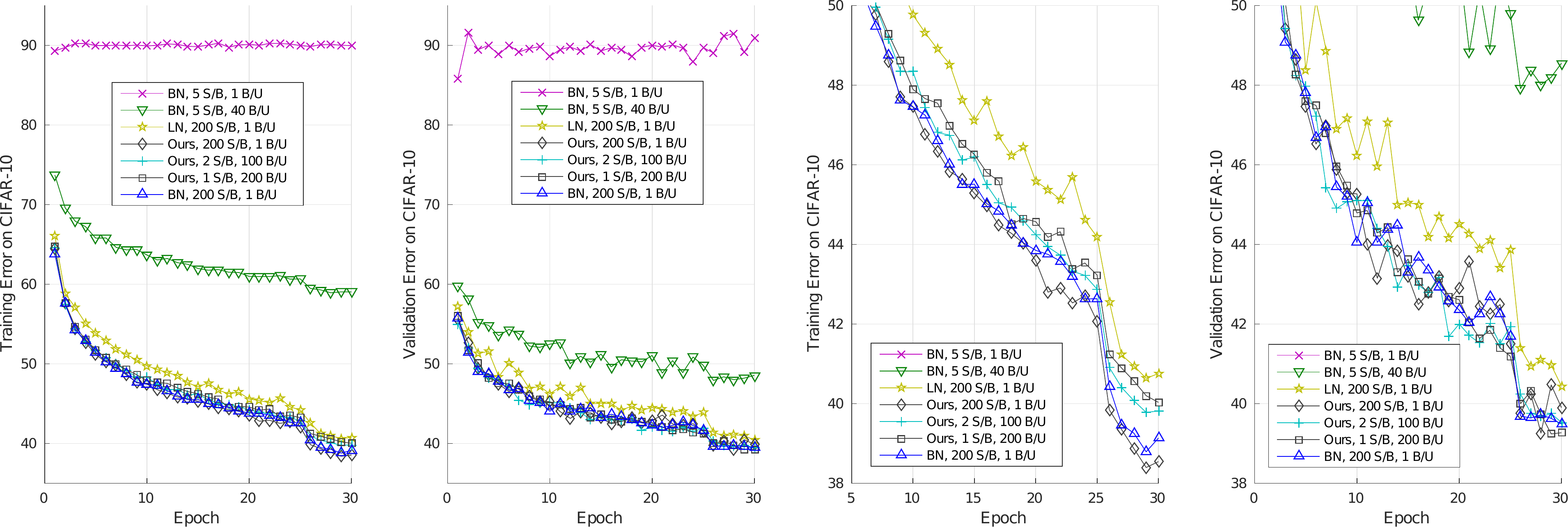}  
  \end{center}
  \caption{Different normalizations applied to a \textbf{feedforward and fully-connected} network (shown in Figure \ref{fig:cifar_arch} A). The right two pannels are \textbf{zoomed-in versions} of the left two pannels.  S/B: Samples per Batch. B/U: Batches per Weight Update.  ``Ours'' refers to Streaming Normalization with ``L1 norm'' (Setting B with p=1 in Section \ref{sec:lp}) and $\alpha_1=\beta_1=0.7$, $\alpha_2=\beta_2=0.3$ and $\beta_3=0$ (see Section \ref{sec:hyper} for more details about hyperparameters). We show that our algorithm works with pure online learning (1 S/B) and tiny mini-batch (2 S/B), and it outperforms Layer Normalization. The choice of S/B does not matter for layer normalization since it processes samples independently. }  
  \label{fig:fc_rest_2}
\end{figure}

\textbf{Layer Normalization vs. Batch Normalization vs. Streaming Normalization:} We compare in Figure \ref{fig:fc_rest_2} Layer Normalization, Batch Normalization and  Streaming Normalization with different choices of S/B and B/U.

\subsection{Evaluating Variants of Batch Normalization}

\begin{figure}[H]
  \renewcommand\figurename{\small Figure}  
  \begin{center}
    \includegraphics[width=\linewidth]{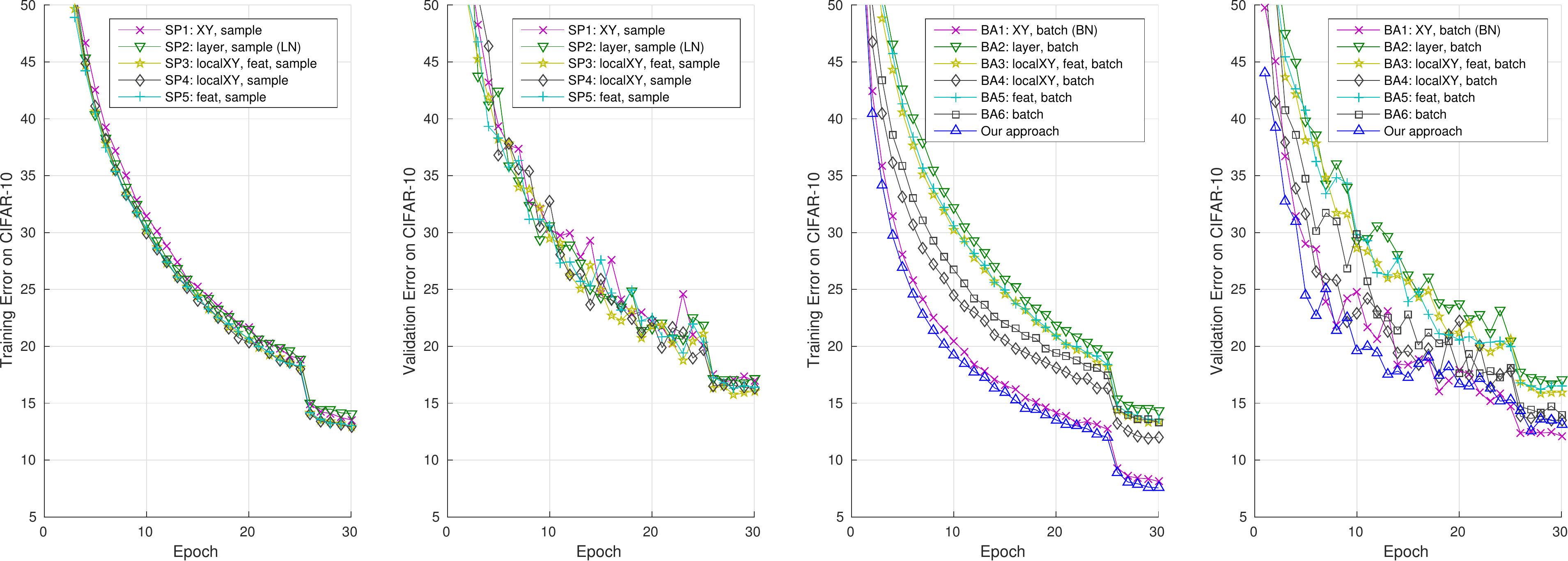} 
  \end{center}
  \caption{Different normalizations applied to a \textbf{feedforward and convolutional} network (shown in Figure \ref{fig:cifar_arch} B). All models were trained with 32 Samples per Batch (S/B), 1 Batch per Update (B/U).  ``Our approach'' refers to Streaming Normalization with ``L2 norm'' (Setting A with p=2 in Section \ref{sec:lp}) and $\alpha_1=\beta_1=0.5$, $\alpha_2=\beta_2=0.5$ and $\beta_3=0$ (see Section \ref{sec:hyper} for more details about hyperparameters). LN: Layer Normalization.  Sample Normalizations (including LN) seem to all work similarly. It seems beneficial to normalize each channel/feature map separately (e.g., compare BA3 with BA4), like what BN does.  }  
  \label{fig:conv_rep_1} 
\end{figure}

\textbf{Feedforward Convolutional Networks:} In Figure \ref{fig:conv_rep_1}, we tested algorithms shown in Figure \ref{fig:general_norm} C and D using the architecture B in Figure \ref{fig:cifar_arch}. We also show the performance of our Streaming Normalization for reference.

\begin{figure}[H]
  \renewcommand\figurename{\small Figure}  
  \begin{center}
    \includegraphics[width=\linewidth]{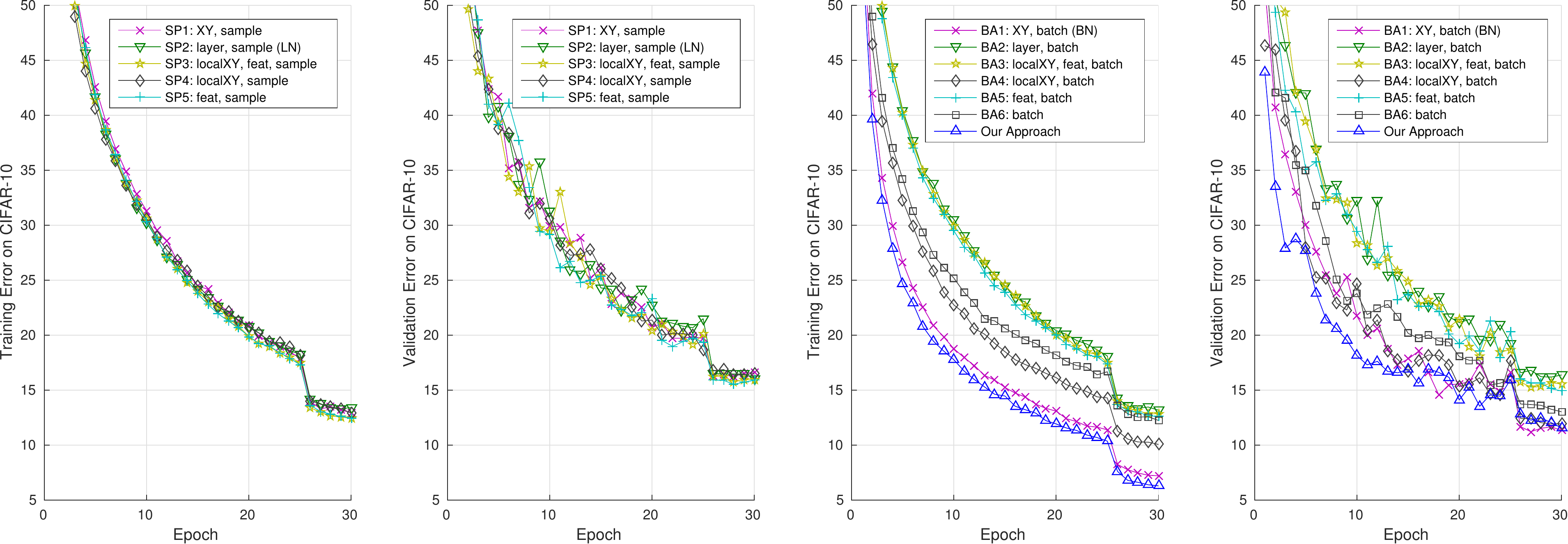} 
  \end{center}
  \caption{Different normalizations applied to a \textbf{recurrent and convolutional} network (Figure \ref{fig:cifar_arch} C with $k_1=5$ and $k_2=1$).  All models were trained with 32 Samples per Batch (S/B), 1 Batch per Update (B/U).  ``Our approach'' refers to Streaming Normalization with ``L2 norm'' (Setting A with p=2 in Section \ref{sec:lp}) and $\alpha_1=\beta_1=0.5$, $\alpha_2=\beta_2=0.5$ and $\beta_3=0$ (see Section \ref{sec:hyper} for more details about hyperparameters). LN: Layer Normalization.  Sample Normalizations (including LN) seem to all work similarly. It seems beneficial to normalize each channel/feature map separately (e.g., compare BA3 with BA4), like what BN does.    }    
  \label{fig:conv_rep_5}
\end{figure}

\textbf{ResNet-like convolutional RNN:} In Figure \ref{fig:conv_rep_5}, we tested algorithms shown in Figure \ref{fig:general_norm} C and D using the architecture C in Figure \ref{fig:cifar_arch}.  We also show the performance of our Streaming Normalization for reference. 

\begin{figure}[H] 
  \renewcommand\figurename{\small Figure}  
  \begin{center}
    \includegraphics[width=\linewidth]{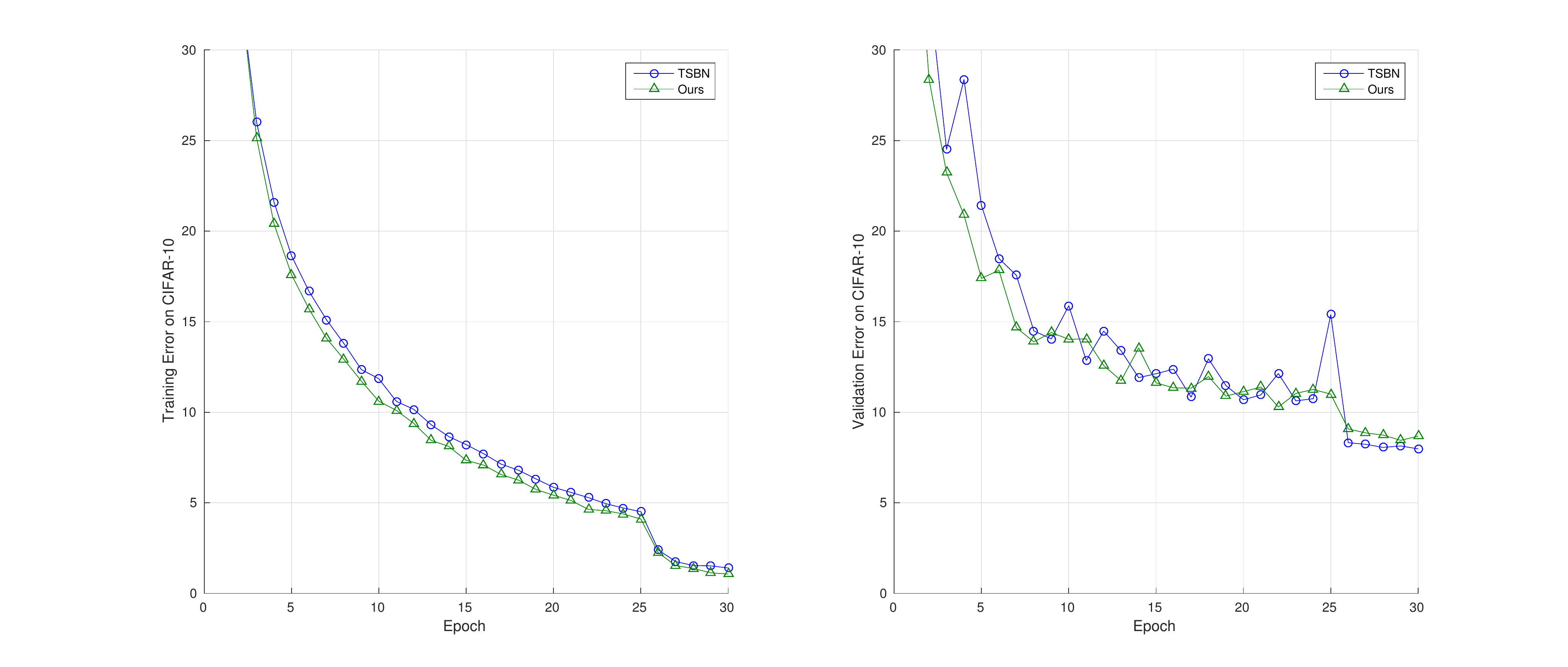}
  \end{center}
  \caption{Time-specific Batch Normalization (TSBN) and Streaming Normalization applied to a \textbf{densely recurrent and convolutional} network (Figure \ref{fig:cifar_arch} D with $k=5$).   ``Ours'' refers to Streaming Normalization with ``L2 norm'' (Setting B with p=2 in Section \ref{sec:lp}), 32 Samples per Batch (S/B) and 2 Batches per Update (B/U) and $\alpha_1=\beta_1=0.7$, $\alpha_2=0.3,\beta_2=0$ and $\beta_3=0.3$ (see Section \ref{sec:hyper} for more details about hyperparameters).  Sometimes for recurrent networks, B/U > 1 is preferred, since the first mini-batch collects NormStats from all timesteps so that the second mini-batch is normalized in a more stable way. TSBN was trained with 64 S/B, 1 B/U (32 S/B, 2 B/U would give similar performance, if not worse). Streaming Normalization has similar performance to TSBN but does not require storing different NormStats for each timestep.}  
  \label{fig:conv_frnn}
\end{figure}

\textbf{Densely Recurrent Convolutional Network:} In Figure \ref{fig:conv_frnn}, we tested Time-Specific Batch Normalization and Streaming Normalization on the architecture D in Figure \ref{fig:cifar_arch}.

\subsection{More Experiments on Streaming Normalization}
\begin{figure}
  \renewcommand\figurename{\small Figure}  
  \begin{center}
    \includegraphics[width=\linewidth]{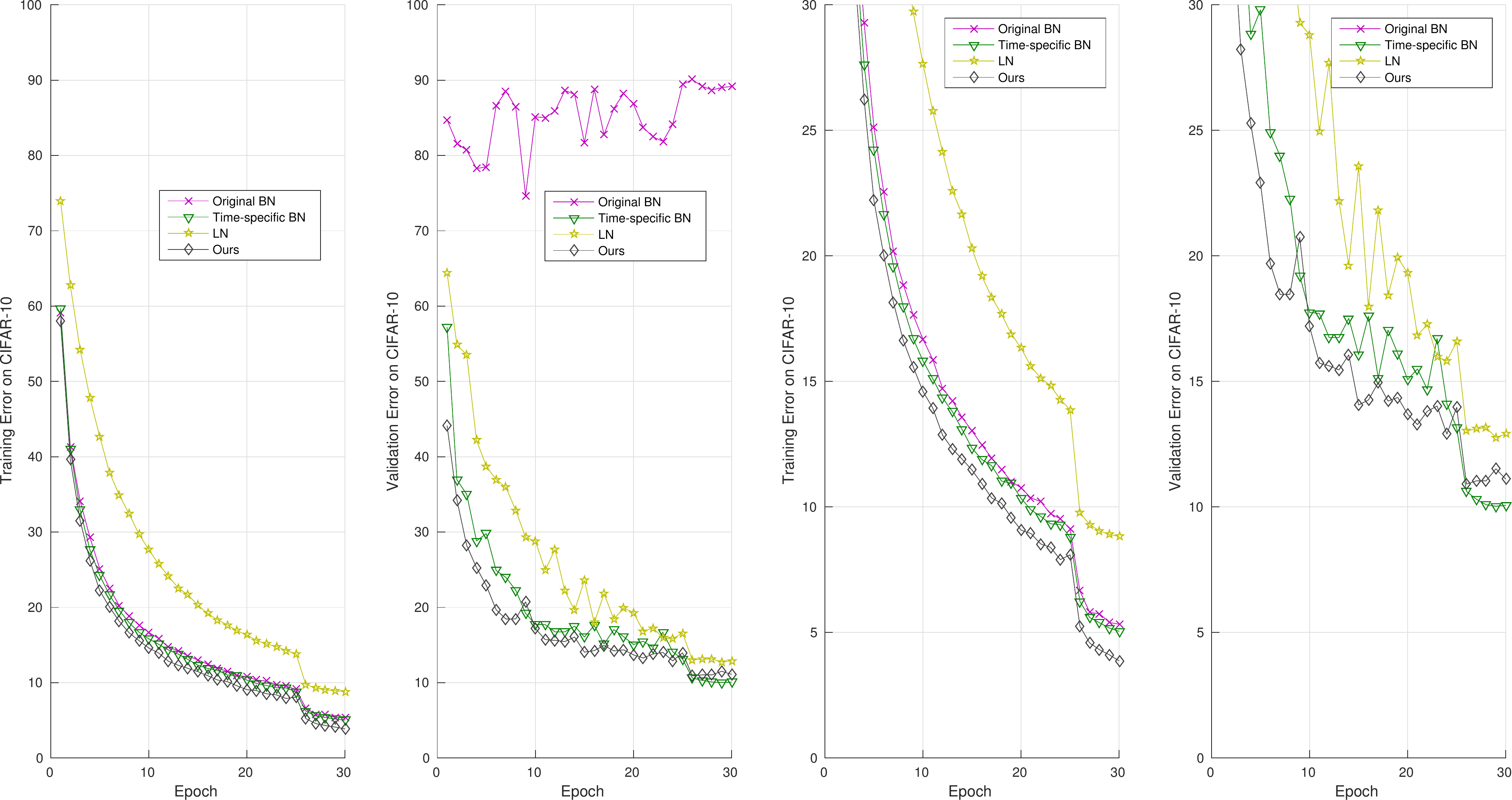} 
  \end{center}
  \caption{Different normalizations applied to a \textbf{recurrent and convolutional} network (shown in Figure \ref{fig:cifar_arch} C with unrolling parameters $k_1=k_2=5$).  The right two pannels are \textbf{zoomed-in versions} of the left two pannels.    ``Ours'' refers to Streaming Normalization with ``L2 norm'' (Setting B with p=2 in Section \ref{sec:lp}), 32 Samples per Batch (S/B), 2 Batches per Update (B/U) and $\alpha_1=\beta_1=0.7$, $\alpha_2=0.3,\beta_2=0$ and $\beta_3=0.3$ (see Section \ref{sec:hyper} for more details about hyperparameters). Time-specific Batch Normalization, original BN and Layer Normalization (LN) were trained with 64 S/B, 1 B/U (32 S/B, 2 B/U would give similar performance, if not worse). Streaming Normalization clearly outperforms other methods in training. Streaming Normalization converges \textbf{more than twice} as fast as LN. Note that 32 S/B 2 B/U and 64 S/B 1 B/U are equivalent to LN since it processes samples independently. Original BN fails on testing. } 
  \label{fig:conv_rep_5_5}
\end{figure}

In Figure \ref{fig:conv_rep_5_5}, we compare the performances of original BN (i.e., NormStats shared over time), time-specific BN, layer normalization and streaming normalization on a recurrent and convolutional network shown in Figure \ref{fig:cifar_arch} C. 

\begin{figure}
  \renewcommand\figurename{\small Figure}  
  \begin{center}
    \includegraphics[width=\linewidth]{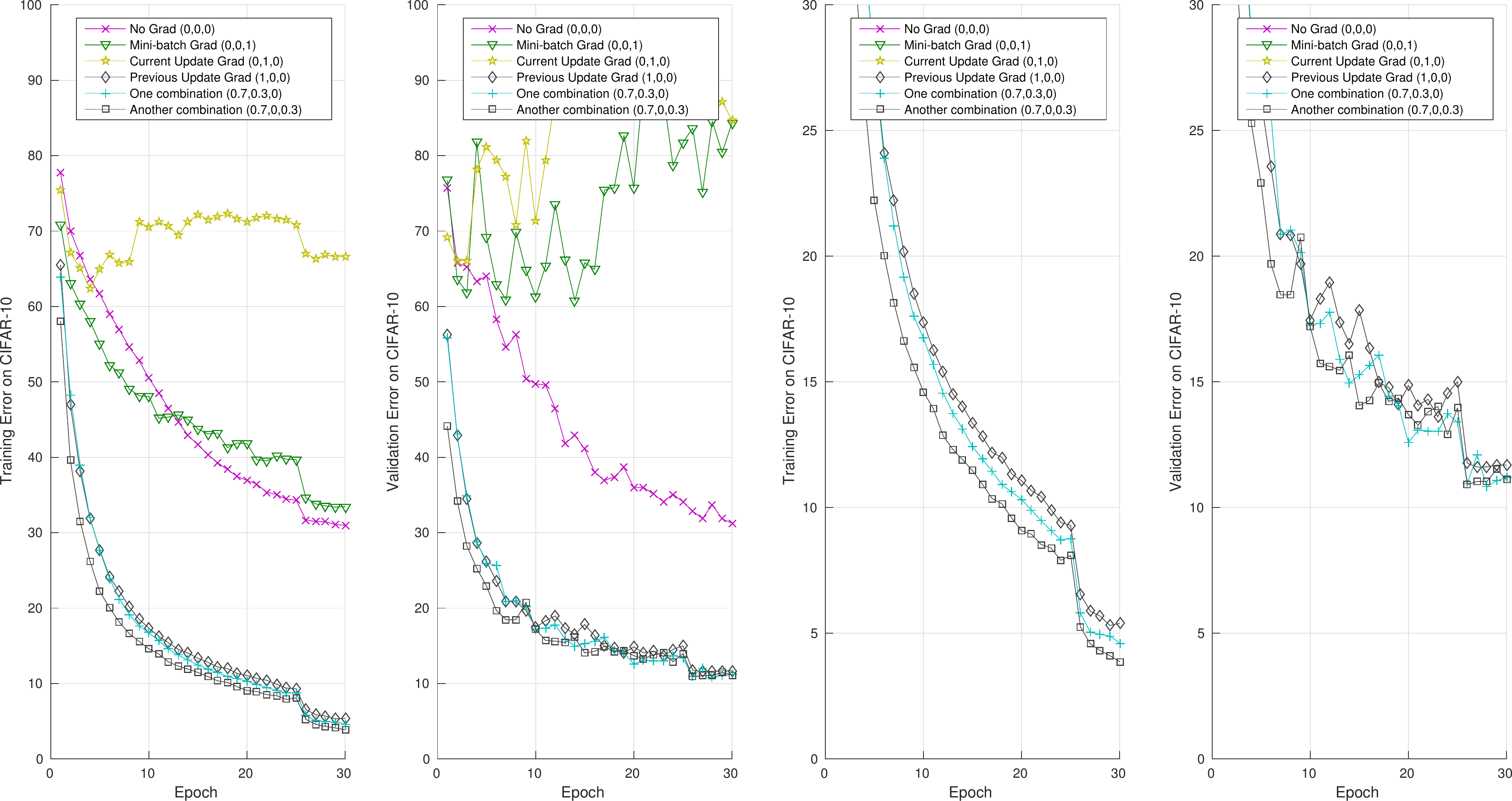} 
  \end{center}
  \caption{Evaluate different choices of hyperparameter  $\beta_1$,$\beta_2$ and $\beta_3$. The architecture is a \textbf{recurrent and convolutional} network (shown in Figure \ref{fig:cifar_arch} C with unrolling parameters $k_1=k_2=5$). The right two pannels are \textbf{zoomed-in versions} of the left two pannels. The models are Streaming Normalization with ``L2 norm'' (Setting B with p=2 in Section \ref{sec:lp}), 32 Samples per Batch (S/B), 2 Batches per Update (B/U) and  $\alpha_1$, $\alpha_2=0.3, \kappa_1=\kappa_3=0.7, \kappa_2=\kappa_4=0.3$. The hyperparameters $(\beta_1,\beta_2,\beta_3)$ are shown in the figure. (0,0,0) means that the gradients of NormStats are ignored. (0,0,1) means only using NormStats gradients from the current mini-batch. (0,1,0) means only using NormStats gradients accumulated since the last weight update. Note that regardless the values of $\beta$, the gradients of NormStats are always accumulated (See Section \ref{sec:stream_grads}). Using gradients from the previous weight update (i.e., 1,0,0) seems to work reasonably well. Some combinations  (i.e., (0.7,0.0.3) or (0.7,0.0.3)) of previous and current gradients seem to give the best performances.  This experiment indicates that streaming the gradients of NormStats is very important for performance. }
  \label{fig:conv_rep_5_5_diff_grad}
\end{figure}

We evaluated different choices of hyperparameter $\beta_1$,$\beta_2$ and $\beta_3$ in Figure \ref{fig:conv_rep_5_5_diff_grad}.

\subsection{Recurrent Neural Networks for Character-level Language Modeling}

\begin{figure}[H]
  \renewcommand\figurename{\small Figure}  
  \begin{center}
    \includegraphics[width=0.9\linewidth]{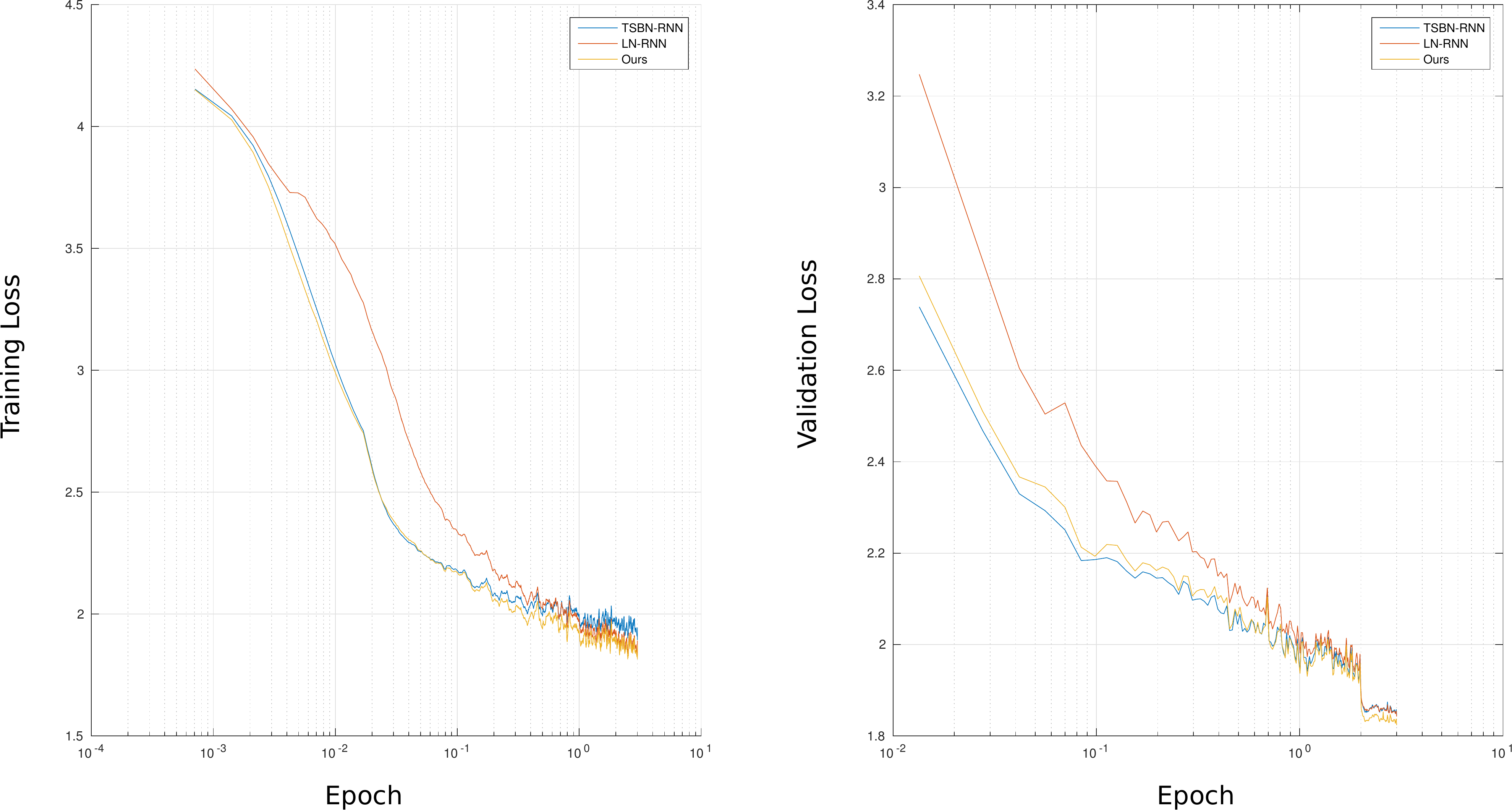} 
  \end{center}
  \caption{Character-level language modeling with RNN on Shakespeare's work concatenated. The training (left) and validation (right) softmax losses are reported.     ``Ours'' refers to Streaming Normalization with ``L2 norm'' (Setting B with p=2 in Section \ref{sec:lp}), 32 Samples per Batch (S/B), 2 Batches per Update (B/U) and $\alpha_1=\beta_1=0.7$, $\alpha_2=0.3,\beta_2=0$ and $\beta_3=0.3$ (see Section \ref{sec:hyper} for more details about hyperparameters).  TSBN: time-specific BN. LN: Layer Normalization. Both TSBN and Streaming Normalization (SN) converges faster than LN. SN reaches slightly lower loss than TSBN and LN. }
  \label{fig:rnn}
\end{figure}

\begin{figure}[H]
  \renewcommand\figurename{\small Figure}  
  \begin{center}
    \includegraphics[width=0.9\linewidth]{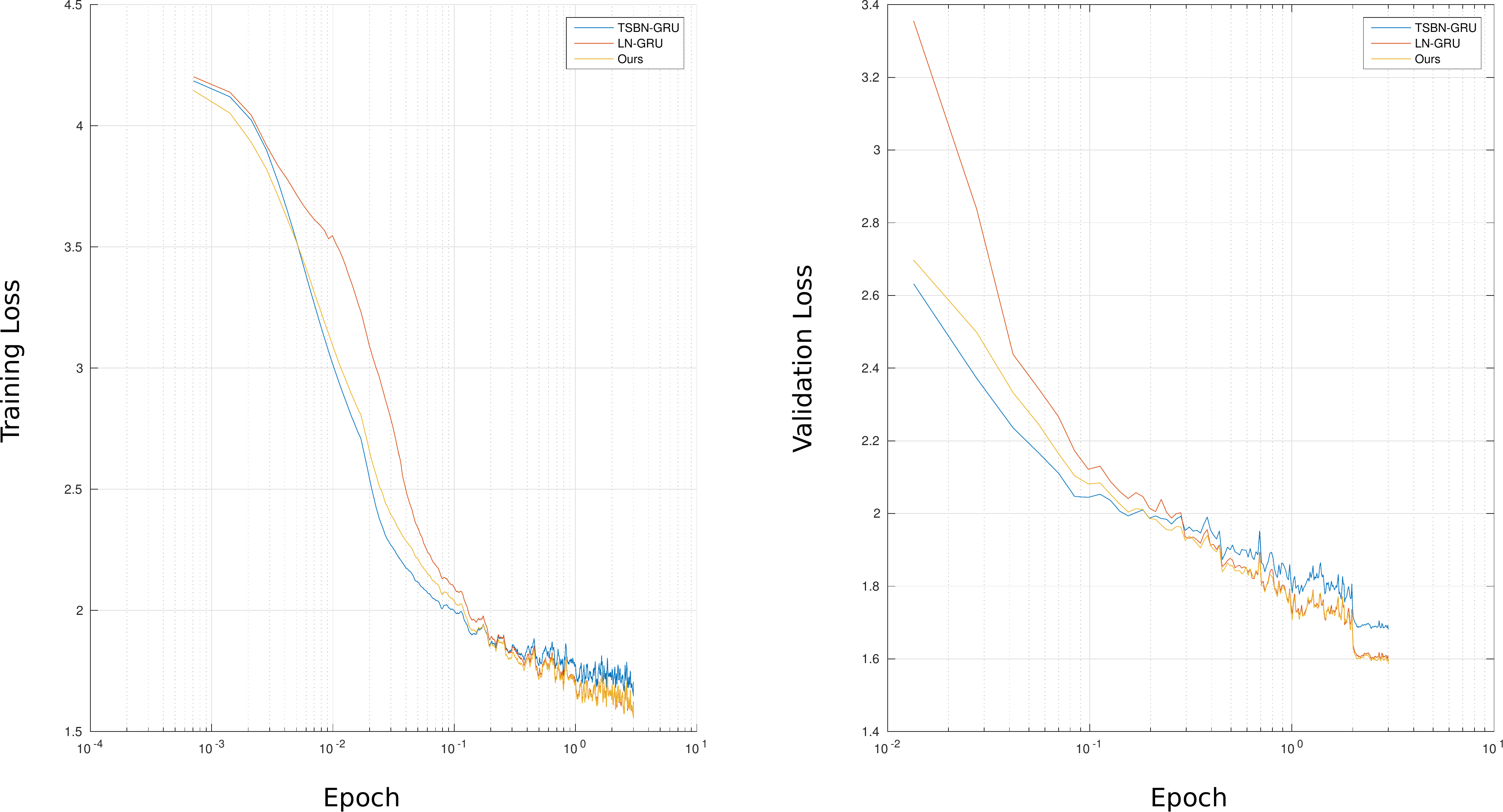}  
  \end{center}
  \caption{Character-level language modeling with GRU on Shakespeare's work concatenated. The training (left) and validation (right) softmax losses are reported.      ``Ours'' refers to Streaming Normalization with ``L2 norm'' (Setting B with p=2 in Section \ref{sec:lp}), 32 Samples per Batch (S/B), 2 Batches per Update (B/U) and $\alpha_1=\beta_1=0.7$, $\alpha_2=0.3,\beta_2=0$ and $\beta_3=0.3$ (see Section \ref{sec:hyper} for more details about hyperparameters).   TSBN: time-specific BN. LN: Layer Normalization. Streaming Normalization converges faster than LN and reaches lower loss than  TSBN.  }     
  \label{fig:gru}
\end{figure}

We tried our simple implementations of vanilla RNN and GRU described in Section \ref{sec:rnn_gru}. The RNN and GRU both have 1 hidden layer with 100 units. Weights are updated using the simple Manhattan update rule described in \cite{liao2015important}. The models were trained with learning rate 0.01 for 2 epochs and 0.001 for 1 epoch on a text file of all Shakespeare's work concatenated. We use 99\% the text file for training and 1\% for validation. The training and validation softmax losses are reported. Training losses are from mini-batches so they are noisy, and we smoothed them using moving averages of 50 neighbors (using the Matlab \textit{smooth} function). The test loss on the entire validation set is evaluated and recorded every 20 mini-batches. We show in Figure \ref{fig:rnn} and  \ref{fig:gru} the performances of Time-specific Batch Normalization, Layer Normalization and Streaming Normalization. Truncated BPTT was performed with 100 timesteps. 

\section{Discussion}

\textbf{Biological Plausibility}


We found that the simple ``Neuron-wise normalization'' (BA6 in Figure
\ref{fig:general_norm} D) performs very well (Figure
\ref{fig:conv_rep_1} and \ref{fig:conv_rep_5}). This setting does not
require collecting normalization statistics from any other neurons. We
show the streaming version of neuron-wise normalization in Figure
\ref{fig:variants}, and the performance is again competitive. In
neuron-wise normalization, each neuron simply maintains running
estimates of its own mean and variance (and related gradients), and
all the information is maintained locally. This approach may serve as
a baseline model for biological homeostatic plasticity mechanisms
(e.g., Synaptic Scaling)
\cite{Turrigiano2004,stellwagen2006synaptic,turrigiano2008self}, where
each neuron internally maintains some normalization/scaling factors
that depend on neuron's firing history and can be applied and updated
in a pure online fashion.



\textbf{Lp Normalization}

Our observations about Lp normalization have several biological implications: First, we show that most Lp normalizations work similarly, which suggests that there might exist a large class of statistics that can be used for normalization. Biological systems could implement any of these methods to get the same level of performance.  Second, L1 normalization is particularly interesting, since its gradient computations are much easier for biological neurons to implement. 


As an orthogonal direction of research, it would also be interesting to study the relations between our Lp normalization (standardizing the average Lp norm of activations) and Lp regularization (discounting the the Lp norm of weights, e.g., L1 weight decay). 


\textbf{Theoretical Understanding}

Although normalization methods have been empirically shown to
significantly improve the performance of deep learning models, there
is not enough theoretical understanding about them.
Activation-normalized neurons behave more similarly to biological
neurons whose activations are constrained into a certain range: is it
a blessing or a curse? Does it affect approximation bounds of shallow
and deep networks \cite{mhaskar2016learning,mhaskar2016deep}? It would
also be interesting to see if certain normalization methods can
mitigate the problems of poor local minima and saddle points, as the
problems have been analysed without normalization \cite{kawaguchi2016deep}.


\textbf{Internal Covariant Shift in Recurrent Networks} 

Note that our approach (and perhaps the brain's ``synaptic scaling'')
does not normalize differently for each timestep. Thus, it does not 
naturally handle internal covariant shift \cite{ioffe2015batch} (more
precisely, covariate shift over time) in recurrent networks, which was
the main motivation of the original Batch Normalization and Layer Normalization.
Our results seem to suggest that internal covariate shift is not as
hazardous as previously believed as long as the entire network's
activations are normalized to a good range. But more research is 
needed to answer this question.



\subsubsection*{Acknowledgments}
This work was supported by the Center for Brains, Minds and Machines (CBMM), funded by NSF STC award CCF – 1231216.

\bibliographystyle{apalike} 

\bibliography{rsdn}

\appendix

\renewcommand{\thefigure}{A\arabic{figure}}

\section{Other Variants of Streaming Normalization} 
\setcounter{figure}{0}  

\begin{figure}[H]
  \renewcommand\figurename{\small Figure}  
  \begin{center}
    \includegraphics[width=\linewidth]{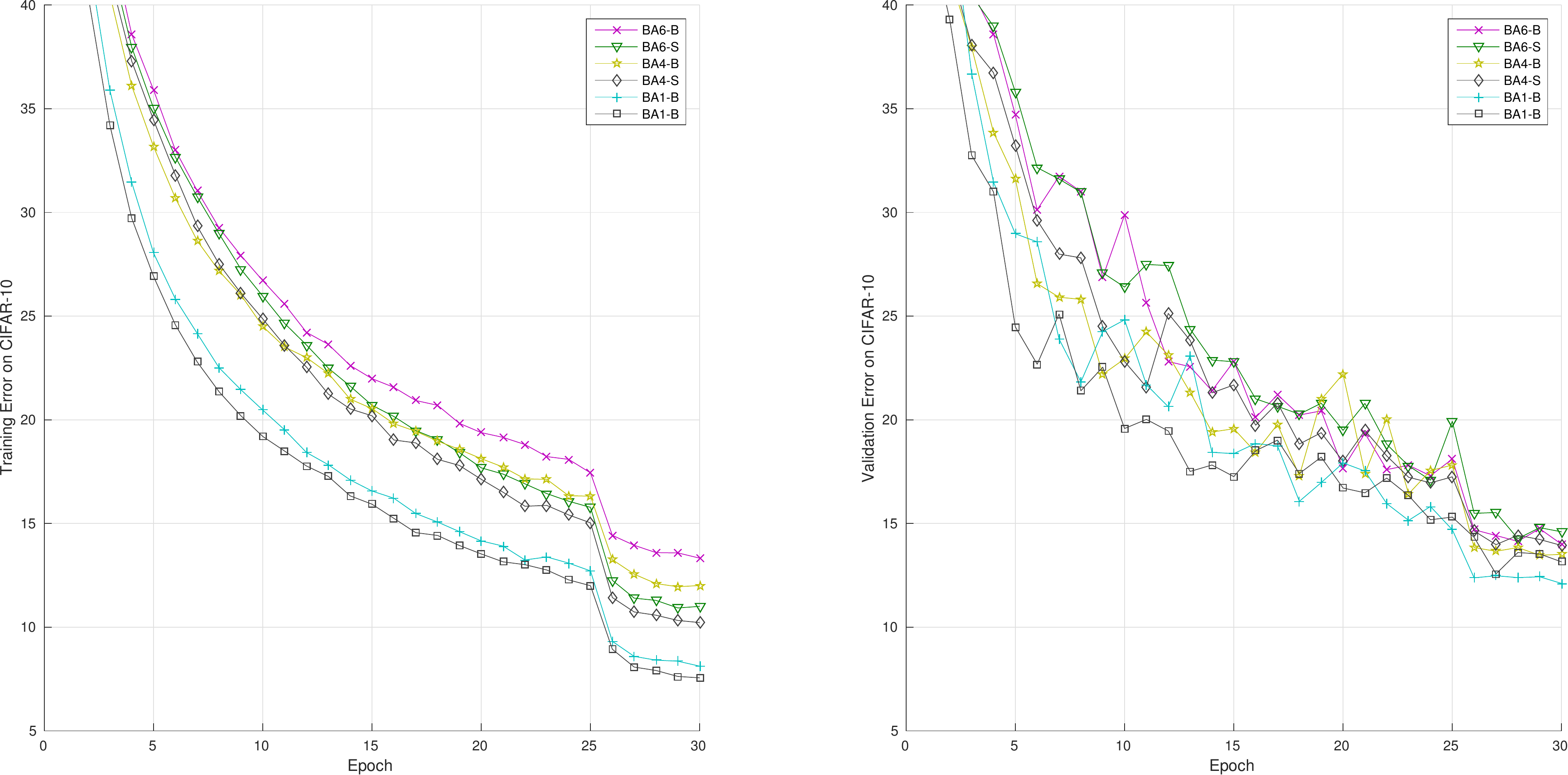}  
  \end{center}
  \caption{We explore other variants of Streaming Normalization with different NormRef (e.g., SP1-SP5, BA1-BA6 in \ref{fig:general_norm} C and D) within each mini-batch. \textbf{-B} denotes the batch version. \textbf{-S} denotes the streaming version.  The architecture is a \textbf{feedforward and convolutional} network (shown in Figure \ref{fig:cifar_arch} B). Streaming significantly lowers training errors.    }  
  \label{fig:variants} 
\end{figure}

\end{document}